\documentclass{article}

\usepackage{microtype}
\usepackage{graphicx}
\usepackage{subcaption}
\usepackage{booktabs} 

\usepackage{hyperref}

\usepackage[preprint]{icml2026}

\usepackage{amsmath}
\usepackage{amssymb}
\usepackage{mathtools}
\usepackage{amsthm}

\usepackage[capitalize,noabbrev]{cleveref}

\theoremstyle{plain}

\theoremstyle{definition}

\theoremstyle{remark}

\usepackage[textsize=tiny]{todonotes}

\usepackage{url}
\usepackage{xurl}
\usepackage{tabularx}
\usepackage{pifont}
\usepackage{xcolor}

\definecolor{markgreen}{RGB}{0,170,0}
\definecolor{xmarkred}{RGB}{220,0,0}
\newcommand{\cmark}{\textcolor{markgreen}{\ding{51}}} 
\newcommand{\xmark}{\textcolor{xmarkred}{\ding{55}}}   

\usepackage{tikz}
\usetikzlibrary{arrows.meta,positioning,calc,shapes.geometric}

\usepackage{setspace}
\usepackage{multicol}
\usepackage{multirow}
\usepackage{tcolorbox}
\usepackage{float}

\usepackage{comment}
\usepackage[dvipsnames]{xcolor}

\icmltitlerunning{FML-bench: Benchmarking Machine Learning Agents for Scientific Research}

\begin{document}

\twocolumn[
  \icmltitle{FML-bench: Benchmarking Machine Learning Agents for Scientific Research}

  \icmlsetsymbol{equal}{*}

  \begin{icmlauthorlist}
    \icmlauthor{Qiran Zou}{equal,nus}
    \icmlauthor{Hou Hei Lam}{equal,nus,thu}
    \icmlauthor{Wenhao Zhao}{nus}
    \icmlauthor{Yiming Tang}{nus}
    \icmlauthor{Tingting Chen}{nus}
    \icmlauthor{Samson Yu}{nus}
    \icmlauthor{Tianyi Zhang}{umn}
    \icmlauthor{Chang Liu}{thu}
    \icmlauthor{Xiangyang Ji}{thu}
    \icmlauthor{Dianbo Liu}{nus}
  \end{icmlauthorlist}

  \icmlaffiliation{nus}{National University of Singapore, Singapore}
  \icmlaffiliation{thu}{Tsinghua University, Beijing, China}
  \icmlaffiliation{umn}{University of Minnesota, Minneapolis, MN, USA}
  
  \icmlcorrespondingauthor{Qiran Zou}{qiranzou@u.nus.edu}
  \icmlcorrespondingauthor{Dianbo Liu}{dianbo@nus.edu.sg}

  \icmlkeywords{Machine Learning, Large Language Models, Research Agents, Benchmarking, Evaluation, Autonomous Research, Exploration Diversity, ICML}

  \vskip 0.3in
]



\printAffiliationsAndNotice{\icmlEqualContribution}

\begin{abstract}

    Large language models (LLMs) have sparked growing interest in machine learning research agents that can autonomously propose ideas and conduct experiments.
    However, existing benchmarks predominantly adopt an engineering-oriented perspective: they emphasize application-oriented tasks and evaluate primarily on final performance and computational cost, overlooking agents' research processes and limiting assessment of their capabilities in scientific research settings.
    To more comprehensively evaluate agents in scientific research settings, we introduce FML-bench, a benchmark comprising 8 diverse and fundamental ML research tasks, and further propose complementary metrics, notably Exploration Diversity, which quantifies the variance of proposals across iterations and reveals how exploration patterns influence research outcomes.
    We evaluate state-of-the-art research agents on FML-bench, showing that agents employing broad exploration strategies exhibit higher exploration diversity and achieve superior performance, and that exploration diversity positively correlates with performance improvements across multiple tasks. We hope these findings and our benchmark inform future agent design and support the community in further investigating agent behavior.
    Our benchmark is available at: \url{https://github.com/qrzou/FML-bench}.
\end{abstract}

\section{Introduction}

Large language models (LLMs) have catalyzed a resurgence of interest in machine learning (ML) research agents which assist or carry out parts of the scientific discovery workflow. 
These agents not only support hypothesis generation, coding, and experiment management, but also increasingly act as collaborators in discovery by providing complementary perspectives that can accelerate machine learning research across domains.
Within this landscape, agents that automatically propose ideas and run experiments are particularly compelling \cite{lu2024ai, yamada2025ai}. 
They close the loop from ideation to empirical validation to maximize automation and to speed up research cycles. 
Compared to settings that only elicit ideas and then use LLMs or humans to assess “novelty” and “feasibility” which often diverge from real-world utility, this approach evaluates agents based on actual experimental outcomes, providing objective and quantitative evidence of their effectiveness \cite{wang2024scimon, baek2024researchagent, si2024can}.

Despite rapid progress, existing benchmarks offer an incomplete picture of research competence, as shown in Tab.~\ref{tab:benchmarks_comparison}. These benchmarks predominantly adopt an engineering-oriented perspective rather than one aligned with scientific research settings, and this limitation manifests in two key aspects. 
\textbf{First}, regarding task construction, most benchmarks focus on Kaggle-style, application-oriented tasks that emphasize engineering execution (e.g., feature engineering, standardized model training, and optimization) while paying limited attention to evaluating an agent's ability to tackle fundamental machine learning research problems, such as representation learning and generalization \cite{chan2024mle, huang2023mlagentbench, padigela2025ml, jing2024dsbench}. 
\textbf{Second}, regarding evaluation design, these benchmarks primarily assess final task performance metrics (e.g., accuracy, recall) and computational cost, while overlooking the characteristics of agents' internal iterative processes, which limits analysis of how agent characteristics relate to research outcomes.

\begin{figure*}[t]
    \centering\includegraphics[width=\linewidth, ]{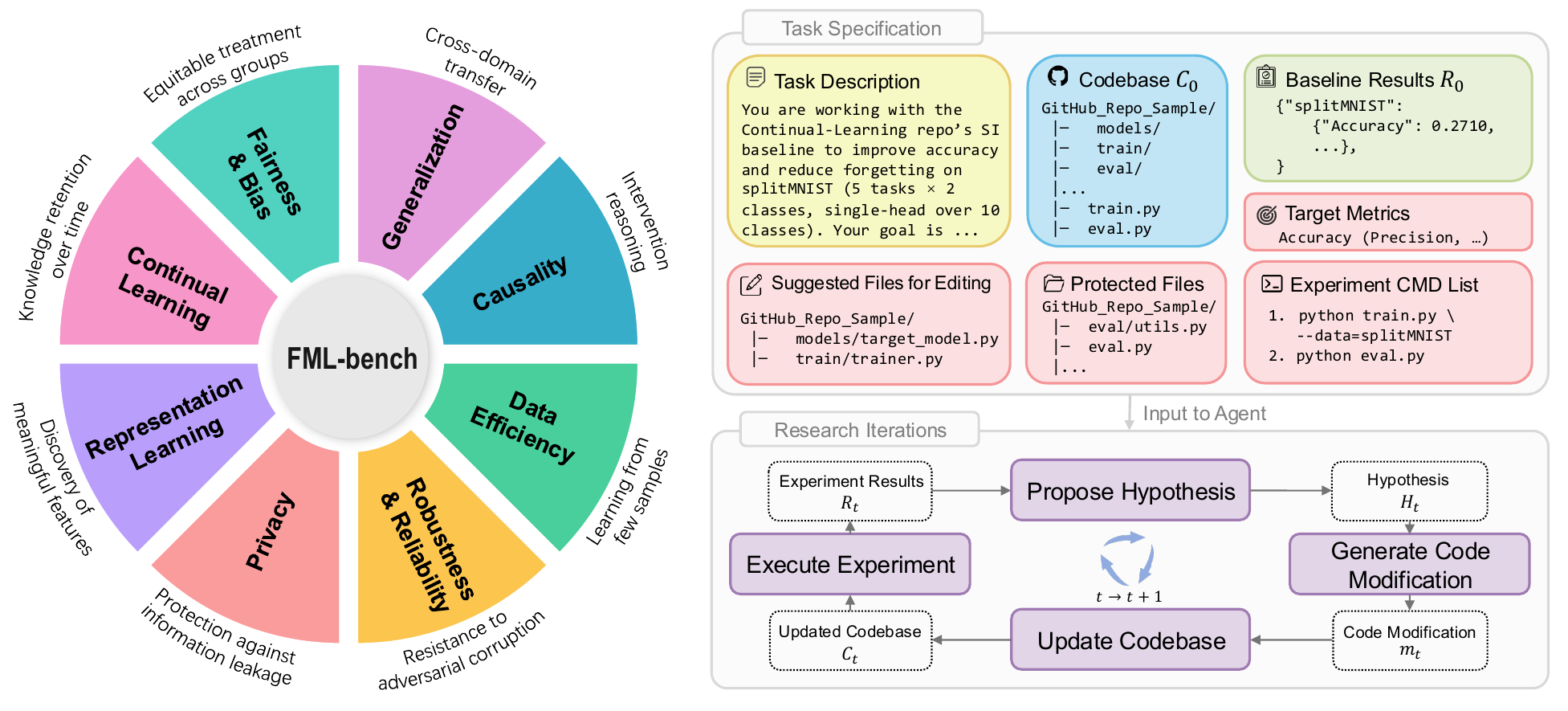}
    \caption{\textbf{Overview of FML-bench.} 
    FML-bench includes 8 fundamental machine learning research tasks, designed to evaluate agents’ capabilities in solving machine learning research problems. 
    Agents are assessed on their ability to solve machine learning problems through iterative research.
}
    \label{fig:benchmark_overview}
    
    \vspace{-1.0em}
    
\end{figure*}

To address these gaps and more comprehensively evaluate agents in scientific research settings, we introduce FML-bench, a benchmark designed to assess automatic ML research agents on fundamental ML problems.
FML-bench comprises 8 diverse tasks (Fig.~\ref{fig:benchmark_overview}) chosen to reflect bottlenecks that repeatedly surface in modern ML. 
The tasks span eight diverse and fundamental ML research problems, including generalization, data efficiency, representation learning, continual learning, causality, robustness, privacy, and fairness (see Section~\ref{sec:benchmark_intro} for task design rationale and references).
Agents are expected to propose new or improved ML methods that deliver stronger empirical results than baselines across these tasks.

Moreover, to enable deeper analysis of agent behavior beyond final outcomes, we further propose complementary metrics that characterize the research process. Notably, we introduce Exploration Diversity, which quantifies the variance of proposals across iterations and reveals how exploration patterns influence research outcomes. Specifically, we compute code embeddings using GraphCodeBERT \cite{guo2020graphcodebert} and measure the dispersion of these embeddings around their centroid. 
Additionally, we propose Step Success Rate and Step Completion Rate to characterize agent reliability throughout the research process. We also report commonly used metrics, including Performance and Cost.

Beyond the two core contributions above, FML-bench addresses additional limitations of existing benchmarks. Some benchmarks provide only raw data without baseline code \cite{chan2024mle, jing2024dsbench}, making it difficult to systematically assess agents' research capabilities while introducing coding barriers that can obscure academic merit (e.g., when sound ideas fail due to engineering pitfalls). Even when baseline codebases are provided, they are often handcrafted and tightly formatted \cite{huang2023mlagentbench, padigela2025ml}. 
In contrast, FML-bench tasks are constructed upon widely recognized ML research repositories with carefully designed task specifications, reflecting how researchers typically build upon established codebases to investigate new ideas.
Furthermore, agents are not required to build entire codebases from scratch but can start from provided baselines, enabling them to focus on scientific advances in algorithms and architectures rather than purely engineering effort.

\begin{table*}[th]
\caption{Comparison of ML agent benchmarks across key design goals.
Repo refers to the repository, and Comp denotes Competition.
$^*$: In MLAgentBench, only part of the tasks meet this requirement; users must prepare baseline and evaluation code even when some tasks are based on real-world Kaggle repositories.}
\label{tab:benchmarks_comparison}
\vspace{-0.2em}
\centering
\begin{tabular}{lccccc}
\toprule
    Design Goals & Ours & MLE--Bench & MLAgentBench & ML--Dev--Bench & DSBench \\
\midrule
    Fundamental ML Problem Focus & \cmark & \xmark & \xmark & \xmark & \xmark \\

    Process-level Evaluation & \cmark & \xmark & \xmark & \xmark & \xmark \\
    
    Real-World Repo/Comp & \cmark & \cmark & ~ \cmark$^*$ & \xmark & \cmark \\
    
    Low Coding Barrier & \cmark & \xmark & \cmark\  & \cmark & \xmark \\

\bottomrule
\vspace{-1.6em}
\end{tabular}
\end{table*}

We evaluate several state-of-the-art automatic research agents on FML-bench. The results show that agents employing broad exploration strategies exhibit higher exploration diversity and achieve superior performance, and that exploration diversity positively correlates with performance improvements across multiple tasks.

We summarize our contributions as follows:

\begin{itemize}

    \item We construct FML-bench, a benchmark for scientific research centered on diverse fundamental ML problems, closing gaps left by application-oriented, engineering-heavy evaluations.

    \item We propose complementary metrics for analyzing agent behavior throughout the scientific research process, notably Exploration Diversity, which quantifies the variance of proposals across iterations, along with Step Success Rate and Step Completion Rate for assessing agent reliability.

    \item We report empirical findings suggesting that broad exploration strategies and high exploration diversity are important contributors to effective ML research agents.
    
\end{itemize}

\section{Related Works}

\subsection{Automatic Research Agents}

With the emergence of large language models (LLMs), research agents have increasingly been explored as tools to support core components of the scientific workflow. These agents are capable of generating and prioritizing research ideas, retrieving and synthesizing literature, and simulating peer review processes.
For instance, SciMON \cite{wang2024scimon} and Nova \cite{hu2024nova} implemented frameworks for generating diverse and novel research ideas. AutoSurvey \cite{wang2024autosurvey} presented an automated literature review framework that performs retrieval over a large arXiv corpus, followed by outline planning and section drafting using specialized models. Meanwhile, AgentReview \cite{jin2024agentreview} employed LLM agents to simulate peer reviews, rebuttals, and committee discussions, offering insights into the dynamics of academic decision-making.

Recent efforts are moving beyond assistance toward fully automatic research agents. These systems aim not only to support researchers but to generate ideas, implement them, run experiments, and refine approaches without human supervision.
One representative system is AIDE \cite{aide2025}, a tree-search agent that optimizes user-defined metrics by iteratively editing and evaluating code, though it executes only one file and modifies a specific target file per iteration. TheAIScientist \cite{lu2024ai} represents an independent line of work, demonstrating end-to-end autonomy across the research process, including idea generation, implementation, experimentation, analysis, and manuscript drafting. Its improved version \cite{yamada2025ai} further reduces reliance on hand-crafted templates, enhancing generality across tasks. Similarly, the AgentLaboratory executes a full pipeline for automatic research, but its evaluation is limited to relatively simple research questions. Separately, AlphaEvolve \cite{novikov2025alphaevolve} adopts an evolutionary approach, iteratively refining and selecting promising ideas through variation and empirical evaluation. Beyond the computer science domain, a growing number of research agents have been developed for other fields, including chemistry, where they are used to investigate and optimize chemical processes \cite{boiko2023autonomous,m2024augmenting}, and biomedical science, where they have been applied to the discovery of novel nanobodies \cite{swanson2024virtual}.

\subsection{Benchmarks for ML Agents}

Existing benchmarks have begun to evaluate agents on code-intensive tasks, yet they remain limited in both scope and flexibility. MLAgentBench \cite{huang2023mlagentbench} includes 13 machine learning engineering tasks, but most are implemented as single-file scripts, which is not practical for real-world scenarios. In addition, it requires to set individual evaluator for each task and lacks support, limiting its scalability to support more tasks. MLE-Bench \cite{chan2024mle} covers 75 Kaggle competitions and assesses whether agents can function as machine learning engineers. It emphasizes tasks such as data pipeline management, experiment orchestration, and submission formatting, which may shift focus away from core machine learning understanding. ML-Dev-Bench \cite{padigela2025ml} places greater emphasis on engineering aspects such as dataset loading and API integration. It evaluates agents' ability to improve existing baselines only in performance tests, which are relatively simple due to narrow task scopes like classification and segmentation, and the use of fixed starter files. In contrast, our benchmark includes tasks spanning diverse machine learning domains. DSBench \cite{jing2024dsbench} aggregates 466 data analysis tasks and 74 modeling tasks from ModelOff and Kaggle, focusing on problem-solving within data science workflows.
By comparison, our benchmark focuses on 8 diverse and fundamental machine learning research tasks. It is built on real-world codebases, thereby providing practical challenges and extensibility by construction, while maintaining a low coding barrier.

\section{Benchmark Task Design}
\label{sec:benchmark_intro}
Prior surveys have identified several critical dimensions that machine learning systems must address to achieve trustworthiness and practical reliability, including robustness, generalization, fairness, and privacy \cite{li2023trustworthy, mehrabi2021survey, wang2022generalizing, goodfellow2014explaining, abadi2016deep}. Complementary work on learning systems emphasizes the importance of data efficiency, representation learning, causality, and continual adaptation for building capable and generalizable AI \cite{adadi2021survey, bengio2013representation, pearl2019seven, chen2018lifelong}. Guided by these foundational challenges, we constructed eight diverse tasks, each representing a distinct research challenge drawn from one of these domains.

For each task, we carefully selected benchmark datasets and baseline methods according to two guiding principles. First, datasets should be both computationally tractable and sufficiently challenging: we required that baseline methods complete training and evaluation within 2 hours, while prioritizing datasets that present meaningful research challenges to differentiate agent capabilities. Second, baseline methods should be classic and widely recognized, yet remain sufficiently below theoretical optima to provide headroom for improvement. This design ensures that agents have sufficient room for exploration while enabling meaningful comparisons of their research capabilities across varying magnitudes of improvement.

\paragraph{Generalization.} This task evaluates the ability to develop algorithms that transfer across domains. We selected ColoredMNIST \cite{arjovsky2019invariant}, which introduces spurious correlations to create a controlled testbed for distribution shift, and adopted ERM \cite{Vapnik1998} as the baseline. Agents are required to propose improved or novel algorithms that enhance out-of-domain accuracy on a held-out target domain while training only on the source domain.

\paragraph{Data Efficiency.} This task measures the ability to enhance few-shot learning under tight data constraints. We selected Mini-ImageNet \cite{vinyals2016matching}, whose fine-grained inter-class similarities present meaningful challenges, and adopted Prototypical Networks \cite{snell2017prototypical} as the baseline. Agents must propose improved metric-based classification algorithms that boost few-shot accuracy while operating under a frozen backbone.

\paragraph{Representation Learning.} This task tests the ability to learn meaningful features from unlabeled data via self-supervised learning. We selected CIFAR-10 \cite{krizhevsky2009learning} for computational efficiency and adopted MoCo \cite{he2020momentum}, a widely recognized contrastive method, as the baseline. Agents are required to improve the pretraining algorithm to achieve higher linear probing accuracy on frozen encoder features.

\paragraph{Continual Learning.} This task evaluates long-term adaptability and the ability to mitigate catastrophic forgetting. We selected splitMNIST \cite{deng2012mnist} under class-incremental learning, where models learn five sequential tasks with a shared output head, and adopted Synaptic Intelligence \cite{zenke2017continual} as the baseline. Agents must propose methods that improve average accuracy across all tasks while avoiding catastrophic forgetting.

\paragraph{Causality.} This task assesses the capacity to estimate treatment effects for intervention reasoning. We selected the IHDP dataset \cite{hill2011bayesian}, a semi-synthetic benchmark enabling ground-truth evaluation, and adopted DragonNet \cite{shi2019adapting} as the baseline. Agents are required to develop improved causal inference strategies that minimize the mean absolute error in treatment effect estimation.

\paragraph{Robustness and Reliability.} This task probes the ability to defend against adversarial data corruption. We constructed a poisoned MNIST dataset with diverse backdoor attacks including edge-based triggers and distributed attack patterns, and adopted dp-instahide \cite{borgnia2021dp} as the baseline defense. Agents must propose defenses that improve the defense score, which balances clean accuracy and resistance to backdoor attacks.

\paragraph{Privacy.} This task assesses the ability to protect against membership inference attacks. We selected CIFAR-10 \cite{krizhevsky2009learning} and adopted Wide-ResNet-28-2 \cite{zagoruyko2016wide} as the baseline, evaluated using both standard and robust membership inference attacks. Agents are required to design defense mechanisms that reduce attack AUC toward 0.5 while maintaining classification accuracy.

\paragraph{Fairness and Bias.} This task measures the ability to balance equitable outcomes with model utility. We selected the COMPAS dataset \cite{angwin2016machine}, which presents documented disparities across protected attributes, and adopted Adversarial Debiasing as the baseline. Agents must minimize absolute average odds difference toward parity while maintaining or improving classification accuracy.

\section{Evaluation Protocol}

\subsection{Evaluation Metrics}
\label{sec:evaluation_metrics}

Existing benchmarks for automatic ML research agents primarily adopt an engineering-oriented evaluation, emphasizing final task performance while overlooking the characteristics of agents’ iterative research processes. To better capture agent behavior throughout scientific discovery, we go beyond standard metrics (Performance and Cost) and introduce additional process-oriented metrics, including Exploration Diversity, Step Success Rate, and Step Completion Rate.

To formally define these metrics, we first describe the iterative research process. Consider an agent conducting research over $T$ iterations. At iteration $t \in \{1, \ldots, T\}$, the agent generates a hypothesis $h_t$, which is instantiated as a code modification $m_t$ applied to the codebase $\mathbf{C}_{t-1}$, yielding $\mathbf{C}_t = \mathbf{C}_{t-1} \oplus m_t$. Let $\mathbf{e}_t$ denote the code embedding of $\mathbf{C}_t$ extracted using GraphCodeBERT~\cite{guo2020graphcodebert}. After completing all iterations, the agent has produced a set of codebases $\{\mathbf{C}_1, \ldots, \mathbf{C}_T\}$ with corresponding embeddings $\{\mathbf{e}_1, \ldots, \mathbf{e}_T\}$. Based on this formulation, we define the following metrics.

\paragraph{Performance.}
Performance serves as the primary objective metric for evaluating agent effectiveness. It measures the best task-specific result achieved across all iterations on the held-out test set. Let $\text{perf}(\mathbf{C}_t)$ denote the evaluation metric for codebase $\mathbf{C}_t$. Performance is defined as:
\begin{equation}
P = \begin{cases}
\max_{t \in \{1, \ldots, T\}} \text{perf}(\mathbf{C}_t) & \text{if higher is better} \\
\min_{t \in \{1, \ldots, T\}} \text{perf}(\mathbf{C}_t) & \text{if lower is better}
\end{cases}.
\end{equation}
This metric captures the agent's ability to discover effective solutions through iterative refinement, evaluated using protected evaluation code to ensure consistent and fair comparison across agents.

\paragraph{Exploration Diversity.}
Exploration Diversity quantifies the variance of proposals across iterations, revealing how exploration patterns influence research outcomes. Let $\bar{\mathbf{e}} = \frac{1}{n} \sum_{t=1}^{n} \mathbf{e}_t$ denote the centroid of all code embeddings, where $n$ is the number of iterations with valid code outputs. Exploration Diversity is computed as:
\begin{equation}
D = \frac{1}{n} \sum_{t=1}^{n} \| \mathbf{e}_t - \bar{\mathbf{e}} \|_2.
\end{equation}
GraphCodeBERT \cite{guo2020graphcodebert} leverages data-flow graphs and code structure rather than surface tokens alone, ensuring that code snippets with identical logic but different variable or function names receive high similarity scores. Such pairs are appropriately treated as low diversity. Greater dispersion indicates broader exploration of implementation strategies within a research run.

Intuitively, higher exploration diversity reflects broader coverage of the hypothesis space, increasing the likelihood of discovering effective solutions. This perspective aligns with classical results in stochastic search and multi-armed bandits~\cite{lai1985asymptotically,bubeck2012regret}, as well as novelty search in evolutionary optimization~\cite{lehman2011abandoning,pugh2016quality}, where broader exploration improves the probability of finding high-reward regions. We verify this relationship empirically in Section~\ref{sec:experiments}.

\paragraph{Step Success Rate.}
Step Success Rate captures the fraction of iterations that produce valid experimental results without errors, reflecting an agent's coding competence and ability to generate syntactically correct, semantically coherent code. It is defined as:
\begin{equation}
S = \frac{N_{\text{suc}}}{N_{\text{comp}}},
\end{equation}
where $N_{\text{suc}}$ is the number of iterations whose experiments execute without errors and yield valid results, and $N_{\text{comp}}$ is the number of iterations actually executed by the agent.

\paragraph{Step Completion Rate.}
Step Completion Rate measures the proportion of executed iterations relative to the assigned total, indicating whether agents can complete full experimental workflows without premature termination. It is defined as:
\begin{equation}
R = \frac{N_{\text{comp}}}{N_{\text{total}}},
\end{equation}
where $N_{\text{total}}$ is the total number of iterations assigned to the agent. Together with Step Success Rate, this metric characterizes overall agent reliability throughout the research process.

\paragraph{Cost.}
Cost accounts for computational and temporal expenditure required during the research process. We report four cost components: (1) total token consumption, the cumulative tokens used across all iterations; (2) step token consumption, the average tokens per iteration; (3) total time consumption, the wall-clock time for a complete experimental run; and (4) step time consumption, the average time per iteration. These components collectively characterize both computational and temporal efficiency.

\subsection{Unified Input-output Interface}

In real-world scientific research, different ML projects often employ distinct codebases with varying execution pipelines, output formats, and evaluation protocols, making it challenging to construct a unified benchmark across diverse research problems. A core design of FML-bench addresses this challenge through unified input-output interfaces that accommodate repository diversity while preserving their inherent complexity. This design enables our benchmark to support diverse ML research tasks built upon existing repositories, bringing evaluation closer to real-world scientific research settings.

\begin{figure*}[t]
    \centering
    \includegraphics[width=\linewidth, ]{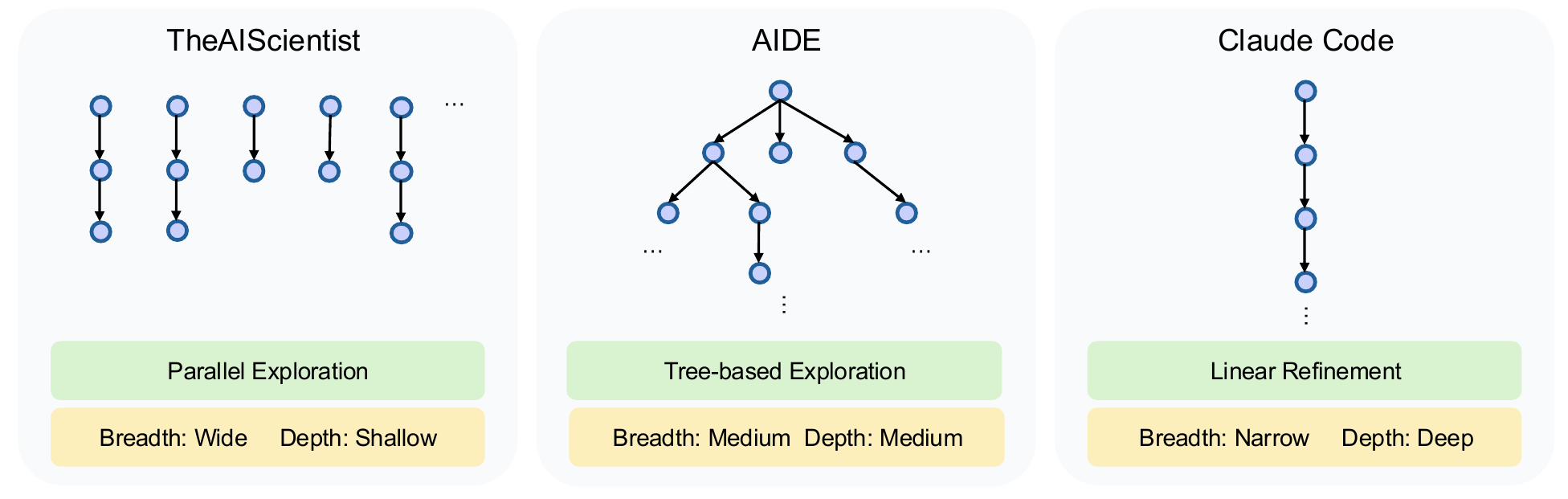}
    \caption{\textbf{Comparison of research exploration strategies of different agents.} TheAIScientist uses parallel exploration for broad coverage, AIDE employs hierarchical tree-based search balancing exploration and exploitation, while Claude Code follows linear refinement for sequential improvement.}
    \label{fig:agent_strategy_comparison}
    \vspace{-0.5em}
\end{figure*}

\paragraph{Input}
Existing benchmark designs struggle to handle different GitHub repositories. Data-based benchmarks accept only datasets and task descriptions as inputs. And benchmarks providing codebase assume unified training script names, single-stage training, no customizable arguments, or requiring to set an individual evaluator manually. In contrast, real repositories use different script names, multi-stage pipelines, diverse training arguments, and include their own evaluator already.
Our solution treats the complete execution sequence (training and evaluation commands) as a single input unit so that the agent receives a command list for running experiments. 
Therefore, our benchmark provide following resources (as shown in Fig. \ref{fig:benchmark_overview}) to agent as input: 1) task description with objectives and expected outputs, 2) complete repository code, 3) suggested files for modification, 4) protected code segments that cannot be modified to preserve evaluation integrity, 5) command list for running experiments, 6) baseline performance, and 7) target improvement metrics.

\paragraph{Output}
Repository outputs are various (e.g. from text files to JSON formats). However, most outputs share a common structure: performance metrics on specific datasets. We provide a post-processing module that converts diverse task outputs into a standardized format, enabling consistent metric extraction across all tasks while preserving native output mechanisms. This design bridges the gap between evaluation standardization and real-world repository diversity, enabling rigorous assessment of agents on practical code optimization tasks.

\paragraph{Evaluation Integrity}
Our benchmark implements protective measures to ensure evaluation integrity. Specifically, we protect evaluation files as read-only and prevent agent modification. Agents must utilize these protected evaluation files to assess their proposed methods.

\section{Experiments}
\label{sec:experiments}

\subsection{Settings}

\paragraph{Selections of Agents}
\label{para:selections_of_agents}

As shown in Fig. \ref{fig:agent_strategy_comparison}, we explore three automatic machine learning research agents, each adopting a distinct research strategy. 
TheAIScientist follows a broad exploration approach, generating and testing a wide range of hypotheses in parallel across multiple experimental directions. 
AIDE employs a hierarchical, tree-based search strategy, balancing the exploration of new possibilities with the exploitation of promising results.
And we prompt Claude Code to employ a linear refinement strategy, sequentially improving its hypotheses and code implementations to address ML tasks.

\paragraph{Experimental Protocol}

Each agent is required to execute in three independent rounds. In each round, the agent is assigned a fixed budget of \textit{total steps} = 100. 
We select the best result achieved among the three rounds based on the target metric computed on the test set.

\subsection{Implementation Details}

\paragraph{Agents}
To enable these agents to operate on our benchmark, several modifications were necessary. We adapted TheAIScientist by fixing compatibility issues and extending its functionality to support the requirements of our benchmark, such as executing experiments in real repositories and reporting results consistently. As for AIDE, we employed its cloud-based commercial variant, Weco, as the operational interface, adapting our benchmark to integrate with its workflow despite limited control over its internal mechanisms. For Claude Code, we designed a prompting scheme (see Appendix~\ref{para:claude_code_prompt}) that enabled it to function as an automatic research agent, capable of reading code, generating hypotheses, and proposing modifications grounded in experimental feedback.

\paragraph{LLMs adopted for agents} 

We employ GPT-5 (2025-08-07) and Gemini-2.5-Pro (2025-06-17) for TheAIScientist and AIDE. For Claude Code, it is constrained to its native models and therefore we use Opus-4.1 (2025-08-05).

\subsection{Results and Discoveries}

\begin{table*}[th]
  \caption{
  \textbf{Comparison of Performance among different agents.}
  G2.5-Pro denotes Gemini-2.5-Pro. 
  $(\uparrow)$ indicates higher is better, $(\downarrow)$ indicates lower is better. 
  }
  \label{tab:overall_performance_comparison}

  \centering
\renewcommand{\arraystretch}{1.08}
\begin{tabular}{l c ccccc}
    \toprule
    
    \multirow{2}{*}{\textbf{ML Problems}} & \multirow{2}{*}{\textbf{Baseline}} & \multicolumn{2}{c}{\textbf{TheAIScientist}} & \multicolumn{2}{c}{\textbf{AIDE}} & \textbf{Claude Code} \\
    \cmidrule(lr){3-4} \cmidrule(lr){5-6} \cmidrule(lr){7-7}
      & & GPT-5 & G2.5-Pro & GPT-5 & G2.5-Pro & Opus-4.1 \\

    \midrule

    Generalization $(\uparrow)$ & 0.2254 & \textbf{0.5036} & 0.3252 & 0.2254 & 0.2254 & \textbf{0.5036} \\
    Data Eff.      $(\uparrow)$ & 0.6547 & 0.7689 & \textbf{0.8231} & 0.6547 & 0.6547 & 0.6571 \\
    Rep. Learn.    $(\uparrow)$ & 0.7562 & 0.7796 & \textbf{0.8597} & 0.8469 & 0.8466 & 0.7725 \\
    Cont. Learn.   $(\uparrow)$ & 0.2710 & 0.4281 & \textbf{0.7808} & 0.4369 & 0.3658 & 0.2337 \\

    Causality    $(\downarrow)$ & 1.1445 & 1.0063 & 0.9925 & 0.9683 & \textbf{0.9549} & 0.9840 \\
    
    Robust. \& Rel.$(\uparrow)$ & 0.4848 & 0.9311 & 0.9205 & 0.9174 & \textbf{0.9633} & 0.8921 \\

    Privacy      $(\downarrow)$ & 0.8114 & 0.4908 & \textbf{0.1750} & 0.4882 & 0.4814 & 0.4892 \\
    
    Fair. \& Bias  $(\downarrow)$ & 0.3787 & 0.0603 & 0.1002 & \textbf{0.0385} & 0.0917 & 0.3787 \\

    \bottomrule
    
\end{tabular}
\end{table*}

\begin{table*}[th]
  \caption{\textbf{Comparisons of different agents across Exploration Diversity, Step Success Rate, Step Completion Rate, Cost.} 
  M stands for Million, H for Hours, and Min for Minutes. 
  $^{*}$ AIDE does not support recording token usage.
  }
  \label{tab:average_comparison_accross_diversity_aca_etc}

  \centering

\begin{tabular}{l ccccc}
    \toprule
    
    \multirow{2}{*}{\textbf{Metrics}} & \multicolumn{2}{c}{\textbf{TheAIScientist}} & \multicolumn{2}{c}{\textbf{AIDE}} & \textbf{Claude Code} \\
    \cmidrule(lr){2-3} \cmidrule(lr){4-5} \cmidrule(lr){6-6}
      & GPT-5 & G2.5-Pro & GPT-5 & G2.5-Pro & Opus-4.1 \\

    \midrule

    Exploration Diversity   & 28.46$^{\pm 9.97}$ 
                            & 20.86$^{\pm 4.42}$ 
                            & 28.60$^{\pm 13.32}$ 
                            & 18.41$^{\pm 12.68}$ 
                            & 8.75$^{\pm 6.07}$ \\

    Step Success Rate $(\uparrow)$ & \textbf{0.83}$^{\pm 0.27}$
        & 0.80$^{\pm 0.26}$
        & 0.64$^{\pm 0.30}$
        & 0.66$^{\pm 0.34}$
        & 0.83$^{\pm 0.32}$ \\

    Step Completion Rate $(\uparrow)$ & \textbf{1.00}$^{\pm 0.00}$
        & \textbf{1.00}$^{\pm 0.00}$
        & 0.54$^{\pm 0.38}$
        & 0.69$^{\pm 0.24}$
        & 0.07$^{\pm 0.06}$ \\

    Total Token Cons. (M) $(\downarrow)$ & 6.04$^{\pm 2.18}$
                        & \textbf{5.45}$^{\pm 2.51}$
                        & N/A$^{*}$
                        & N/A$^{*}$
                        & 9.06$^{\pm 3.64}$ \\

    Step Token Cons. (M) $(\downarrow)$ & 0.06$^{\pm 0.02}$
                        & \textbf{0.05}$^{\pm 0.03}$
                        & N/A$^{*}$
                        & N/A$^{*}$
                        & 3.10$^{\pm 4.34}$ \\

    Total Time Cons. (H) $(\downarrow)$  & 9.42$^{\pm 5.25}$
                        & 11.74$^{\pm 8.37}$
                        & 5.46$^{\pm 6.92}$
                        & 9.49$^{\pm 7.24}$
                        & \textbf{1.18}$^{\pm 1.80}$ \\

    Step Time Cons. (Min) $(\downarrow)$  & 11.05$^{\pm 10.71}$
                        & 14.23$^{\pm 14.55}$
                        & 12.55$^{\pm 12.17}$
                        & \textbf{10.48}$^{\pm 8.29}$
                        & 13.26$^{\pm 11.29}$ \\

    \bottomrule
\end{tabular}
\end{table*}

\subsubsection{Comparison of Agents with Different Research Strategies}
\label{sec:comparison_search_strategies}
As shown in Tab.~\ref{tab:overall_performance_comparison}, the combination of TheAIScientist with Gemini-2.5-Pro achieved the best performance, ranking first in 4 out of 8 tasks. The combination of AIDE with Gemini-2.5-Pro ranked second, securing top results in 2 out of the 8 tasks.
These findings suggest that TheAIScientist performs better in discovering novel and effective machine learning methods, compared to AIDE and Claude Code.

As illustrated in Fig.~\ref{fig:agent_strategy_comparison}, TheAIScientist adopts a research exploration strategy that is broad but shallow, while AIDE maintains both medium breadth and depth. In contrast, Claude Code exhibits a narrow yet deep exploration pattern. 
For a detailed explanation of the research strategies adopted by each agent, refer to Section~\ref{para:selections_of_agents}.
Considering the research exploration strategy together, the results suggest that a broader research exploration space adopted by TheAIScientist is more effective for discovering promising ideas.
This insight offers practical guidance for machine learning research agent that
broadly exploring diverse ideas could be more productive than focusing on a single direction.

\subsubsection{Analysis of Exploration Diversity}

As shown in Tab.~\ref{tab:overall_performance_comparison} and Tab.~\ref{tab:average_comparison_accross_diversity_aca_etc}, several key findings emerge from our experimental results. First, TheAIScientist and AIDE demonstrate significantly higher exploration diversity compared to Claude Code, while also exhibiting superior overall performance across the tasks. Second, when both systems employ GPT-5 as the underlying language model, TheAIScientist and AIDE achieve comparable diversity scores (28.46 vs. 28.60, respectively), and both methods attain one best result each. Third, when utilizing Gemini-2.5-Pro, TheAIScientist exhibits higher diversity than AIDE (20.85 vs. 18.41) and achieves a greater number of best results (4 vs. 2). 
These observations suggest that, with the LLM held fixed, agents with greater exploration diversity tend to achieve better outcome performance. In contrast, when the agent is held fixed, exploration diversity is not a decisive factor; performance also depends on other capabilities of different LLMs (e.g., reasoning and code-writing ability).

Furthermore, TheAIScientist achieves a mean exploration diversity of 24.66, which exceeds both AIDE's 23.50 and Claude Code's 8.75. This enhanced diversity can potentially be attributed to TheAIScientist's more exploration breadth-oriented search strategy, as illustrated in Sec.~\ref{sec:comparison_search_strategies}.
\begin{table*}[t] %
    \centering
    \vspace{0.5em}
    \caption{\textbf{Correlation analysis between exploration diversity and outcome.} When computing the correlation, the outcome scores are negated for tasks where lower values indicate better performance, ensuring that positive correlations consistently represent improved performance with increased exploration diversity.}
    \label{tab:correlation_tab}
    %
    \begin{tabular}{lcccccccc}
        \toprule
        Metric & Gen. & Data Eff. & Rep. Learn. & Cont. Learn. & Causality & Robust. & Privacy & Fair. \\
        \midrule
        p-value & 0.1137 & 0.0121 & 0.1358 & 0.0002 & 0.4668 & 0.1944 & 0.6479 & 0.0517 \\
        Correlation & \textbf{0.4418} & \textbf{0.6287} & \textbf{0.4036} & \textbf{0.8374} & -0.2036 & -0.3846 & -0.1286 & \textbf{0.5292} \\
        \bottomrule
    \end{tabular}
    \vspace{0.5em}
\end{table*}

To further validate the relationship between exploration diversity and performance, we computed the correlation coefficients between exploration diversity and performance. As presented in Tab.~\ref{tab:correlation_tab}, our findings reveal positive correlations across five tasks, with p-values indicating significant, borderline significant, or suggestive evidence of this relationship ($p < 0.05$ for significant; $p < 0.10$ for borderline significant; $p < 0.15$ for suggestive evidence; $p \geq 0.15$ for not significant). Notably, for these five tasks, higher exploration diversity was associated with better performance.

\subsubsection{Token and Time Consumption}
Tab.~\ref{tab:average_comparison_accross_diversity_aca_etc} reports both the average token and time consumption per step, as well as the total token and time usage for a complete experimental run.
We observe that TheAIScientist consumes more tokens than AIDE, while Claude Code, despite its lower performance, uses the highest number of tokens among the three agents.
This indicates that dedicated automatic ML research agents, such as TheAIScientist and AIDE, are more suitable for ML research problems in terms of both performance and token efficiency compared to general-purpose agents like Claude Code.
In terms of time per step, all three agents show similar durations, with differences of around 2 minutes, which are not substantial.
However, AIDE and Claude Code exhibit significantly shorter total execution times for a full experiment. This is primarily due to premature termination issues, as evidenced by the step completion rate, which leads to reduced overall time usage.

\subsubsection{Case Analysis}
\paragraph{Worst Case Analysis}
TheAIScientist strongly favors algorithmic modifications (56.2\%) with zero implementation errors; AIDE presents balanced distribution but higher implementation issues (31.2\%); while Claude Code shows elevated parameter configuration rates (37.5\%) matching its algorithmic modification rate. Across LLMs, Gemini-2.5-Pro exhibits extremely strong algorithmic modification tendency (68.8\%), GPT-5 primarily focuses on structural modifications (50.0\%), and Opus-4.1 maintains balanced distribution between parameter configuration and algorithmic modifications. See Appendix~\ref{app:case_analysis} for detailed results.

\paragraph{Best Case Analysis}
 TheAIScientist shows a strong algorithmic innovation focus, predominantly targeting core algorithmic logic; AIDE combines strong algorithmic innovation with hyperparameter tuning; while Claude Code uniquely favors architectural modifications including layer additions and normalization adjustments. Across LLMs, GPT-5 exclusively prioritizes algorithmic modifications (100.0\%), Gemini-2.5-Pro exhibits the most diversified approach with unique data augmentation efforts (28.6\%), and Opus-4.1 balances attention between architectural and algorithmic modifications. See Appendix~\ref{app:case_analysis} for detailed results.

\subsubsection{Additional Observations}

\paragraph{Shallow edits}
We found that AIDE sometimes misinterprets the structure and logic of target codebases. In certain cases, it generated new classes or components that were never integrated into the actual execution, resulting in no functional improvement over the baseline.
As shown in Tab.~\ref{tab:overall_performance_comparison}, AIDE failed to improve the baseline in tasks related to Generalization and Data Efficiency.
This may stem from the fact that AIDE only supports iterative modifications on a single file. However, real-world ML research codebases are often complex and span multiple files, making AIDE insufficient for addressing realistic research tasks.

\paragraph{Premature termination}
We encountered the issue of early termination of AIDE and Claude Code. For AIDE, the agent sometimes terminated prematurely due to its commercial version Weco which relies on cloud infrastructure that occasionally failed during execution. For Claude Code, early stopping was often triggered by the model's internal reasoning, where the LLM would decide not to continue even when further actions are possible.

\section{Conclusion}

In this work, we introduce FML-bench, a benchmark designed to evaluate ML research agents in scientific research settings. Unlike existing benchmarks that adopt an engineering-oriented perspective through application-oriented tasks and performance-focused evaluation, FML-bench comprises 8 diverse and fundamental ML research problems constructed upon widely recognized academic repositories. We further propose complementary metrics that characterize agent behavior throughout the research process, notably Exploration Diversity, along with Step Success Rate and Step Completion Rate for assessing agent reliability.
Through systematic evaluation of state-of-the-art research agents on FML-bench, we find that agents employing broad exploration strategies exhibit higher exploration diversity and achieve superior performance, and that exploration diversity positively correlates with performance improvements across multiple tasks. We hope FML-bench and our findings provide a foundation for evaluating research agents from a scientific research perspective and offer insights for building more effective research agents.

\section*{Impact Statement}
This work impacts ML applications by shifting the evaluation of research agents from narrow, answer-matching benchmarks toward realistic, process-oriented scientific problem solving, which may lead to agents that generalize better and provide more reliable assistance in real research workflows. While research agents themselves may raise ethical concerns (e.g., misuse or over-automation of scientific judgment), FML-bench does not introduce inherent ethical risks; rather, it serves as a neutral tool for observing and analyzing agent behavior. 
Moreover, by grounding tasks in realistic research problems and emphasizing the investigation of agent behavior, FML-bench promotes more meaningful progress and greater transparency in research agent evaluation. 
In addition, our proposed unified input–output interfaces enable extensibility across repositories, datasets, and baselines, which we believe can help encourage diverse, reproducible, and behavior-focused evaluation.

\section*{Acknowledgements}
We gratefully acknowledge Zhengyao Jiang and Weco (\url{https://www.weco.ai/})
for their support and for providing access to their more general agent, which extended beyond the limitations of the original AIDE and enabled us to run AIDE as a baseline on our benchmark.


\bibliography{example_paper}
\bibliographystyle{icml2026}

\newpage
\appendix
\onecolumn
\section{Task Design Details}
\label{sec:app_task_settings}


\paragraph{Generalization.} This task evaluates the ability to develop algorithms that transfer effectively across domains, which is critical for robust real-world deployment. We selected the ColoredMNIST dataset \cite{arjovsky2019invariant}, which introduces spurious correlations between color and class labels, creating a controlled yet challenging testbed for distribution shift. As the baseline, we adopted Empirical Risk Minimization (ERM) \cite{Vapnik1998}, the most fundamental training paradigm that provides a clear reference point for measuring generalization improvements. We constructed this task using the DomainBed repository \cite{gulrajani2020search}, a widely used framework for domain generalization research. Agents are required to train on a source domain and are evaluated on a held-out target domain under distribution shift. The evaluation metric is out-of-domain accuracy, with the objective of maximizing generalization performance while maintaining in-domain accuracy.

\paragraph{Data Efficiency.} This task measures the ability to enhance few-shot learning performance by optimizing decision rules under tight data constraints. We selected the Mini-ImageNet dataset \cite{vinyals2016matching}, a standard few-shot learning benchmark whose fine-grained inter-class similarities and limited training examples per class present meaningful challenges for metric-based classification. As the baseline, we adopted Prototypical Networks \cite{snell2017prototypical}, a classic and widely recognized metric-based approach that computes class prototypes in the embedding space. We constructed this task using the Easy-Few-Shot-Learning repository \cite{SicaraEasyFSL}. Agents must operate under a frozen backbone and propose improved algorithms for metric-based classification in the embedding space. The evaluation metric is few-shot classification accuracy, with the objective of improving the classifier's ability to generalize to novel classes.

\paragraph{Representation Learning.} This task tests the ability to learn generalizable and semantically meaningful features from unlabeled data through self-supervised learning. We selected the CIFAR-10 dataset \cite{krizhevsky2009learning}, which provides a computationally efficient yet sufficiently challenging benchmark for evaluating learned representations. As the baseline, we adopted MoCo (Momentum Contrast) \cite{he2020momentum}, a widely recognized and effective contrastive learning method that maintains a momentum-updated encoder for consistent representations. We constructed this task using the Lightly repository \cite{lightly}, a modular framework for self-supervised learning. Agents pretrain encoders on the CIFAR-10 training split without labels using a ResNet-18 backbone, and evaluation is conducted via linear probing accuracy with the encoder frozen. The objective is to improve representation quality such that a linear classifier trained on frozen features achieves higher Top-1 accuracy on the test split, while keeping parameter count and FLOPs comparable to the baseline.

\paragraph{Continual Learning.} This task evaluates long-term adaptability in non-stationary environments and the ability to mitigate catastrophic forgetting when learning sequentially across multiple tasks. We selected the splitMNIST dataset \cite{deng2012mnist} under the class-incremental learning scenario, where the model must learn five sequential tasks (each introducing two new digit classes) using a shared ten-way classification head—a setting that rigorously tests knowledge retention. As the baseline, we adopted Synaptic Intelligence (SI) \cite{zenke2017continual}, an established regularization-based method that estimates parameter importance online to protect previously learned knowledge. We constructed this task using the Continual-Learning repository \cite{vandeven2022three}. The evaluation metric is average accuracy across all five tasks, with the objective of maximizing performance while avoiding unfair advantages in model size or computational budget.

\paragraph{Causality.} This task assesses the capacity to reason about interventions and estimate treatment effects, which is essential for decision-making in high-stakes environments. We selected the IHDP (Infant Health and Development Program) dataset \cite{hill2011bayesian}, a semi-synthetic benchmark built from a real randomized controlled trial with simulated treatment effects, enabling ground-truth evaluation of causal inference methods. As the baseline, we adopted DragonNet \cite{shi2019adapting}, an established neural network architecture designed for treatment effect estimation that jointly models outcome prediction and propensity scoring. We constructed this task using the CausalML repository \cite{chen2020causalml}. The evaluation metric is mean absolute error in estimating individual treatment effects, with the objective of improving the precision of causal effect estimation.

\paragraph{Robustness and Reliability.} This task probes the ability to improve model robustness against adversarial data corruption while preserving performance on clean data. We constructed a poisoned MNIST dataset incorporating diverse backdoor attack methods, including edge-based triggers and various distributed attack patterns, to create a challenging defense scenario. As the baseline, we adopted dp-instahide \cite{borgnia2021dp}, a privacy-preserving defense that mixes inputs with public data and applies differential privacy noise to hinder inversion and poisoning attacks. We constructed this task using the Adversarial Robustness Toolbox (ART) repository \cite{nicolae2018adversarial}. Agents are tasked with proposing defenses that reduce the effectiveness of poisoning attacks while maintaining high clean accuracy. The evaluation metric is a defense score defined as the harmonic mean of clean accuracy and resistance accuracy against backdoor attacks.

\paragraph{Privacy.} This task is critical for assessing the ability to enhance privacy protections against membership inference attacks, which attempt to determine whether specific data points were used in model training. We selected the CIFAR-10 dataset \cite{krizhevsky2009learning} as the training corpus, providing a standard image classification setting for privacy evaluation. As the baseline, we adopted Wide-ResNet-28-2 \cite{zagoruyko2016wide}, a widely used architecture that provides a representative target model for privacy auditing. We constructed this task using the PrivacyMeter repository \cite{murakonda2020ml}, which implements both standard Membership Inference Attacks (MIA) and Robust MIA (RMIA) that leverages reference models and population data for more rigorous privacy assessment. The evaluation metric is the attack AUC, with the objective of driving it toward 0.5 (random guessing) while maintaining classification accuracy, thereby reducing information leakage about training data membership.

\paragraph{Fairness and Bias.} This task measures the ability to balance equitable outcomes across demographic groups with overall model utility in classification settings. We selected the COMPAS dataset \cite{angwin2016machine}, a recidivism prediction benchmark that presents meaningful fairness challenges due to documented disparities across protected attributes such as race and sex. As the baseline, we adopted Adversarial Debiasing, which employs a classifier-adversary architecture to learn fair representations by preventing the adversary from predicting sensitive attributes. We constructed this task using the AIF360 repository \cite{aif360-oct-2018}. The evaluation metric is absolute average odds difference, with the objective of minimizing this disparity toward parity (zero) while maintaining or improving classification accuracy across protected subgroups.

\section{Case Analysis}
\label{app:case_analysis}
\subsection{Worst Case Analysis}

\paragraph{Modification Distribution by Agents} 
As shown in Tab.~\ref{tab:mod_dist_worst_case_agents}, TheAIScientist strongly favors algorithmic modifications (56.2\%) with zero implementation errors, indicating robust execution. AIDE presents a more balanced modification distribution but experiences higher implementation and execution issues (31.2\%). Claude Code shows elevated parameter configuration rates (37.5\%), equivalent to its algorithmic modification rate.

\paragraph{Modification Distribution by LLMs} 
As shown in Tab.~\ref{tab:mod_dist_worst_case_llms}, Gemini-2.5-Pro strongly favors algorithmic modifications (68.8\%), preferring core algorithm changes; GPT-5 primarily focuses on structural modifications (50.0\%) with emphasis on architecture adjustments; while Opus-4.1 maintains balanced distribution between parameter configuration and algorithmic modifications.

\subsection{Best Case Analysis}

\paragraph{Modification Distribution by Agents} 
As shown in Tab.~\ref{tab:mod_dist_best_case_agents}, TheAIScientist shows strong algorithmic innovation focus, predominantly targeting core algorithmic logic; AIDE combines strong algorithmic innovation with hyperparameter tuning; while Claude Code uniquely favors architectural modifications, including layer additions and normalization adjustments.

\paragraph{Modification Distribution by LLMs} 
As shown in Tab.~\ref{tab:mod_dist_best_case_llms}, GPT-5 exclusively prioritizes algorithmic modifications (100.0\%); Gemini-2.5-Pro exhibits the most diversified approach, uniquely incorporating substantial data augmentation efforts (28.6\%); while Opus-4.1 balances attention between architectural and algorithmic modifications.

\begin{table}[th]
    \centering
    \caption{Worst Case Modification Distribution by Agents}
    \label{tab:mod_dist_worst_case_agents}
    \begin{tabular}{lccc}
        \toprule
        Modification Type & TheAIScientist & AIDE & Claude Code \\
        \midrule
        Algorithmic Modification    & 56.2\% (9) & 37.5\% (6) & 37.5\% (3) \\
        Structural Modification     & 31.2\% (5) & 25.0\% (4) & 12.5\% (1) \\
        Parameter Configuration     & 12.5\% (2) & 6.2\% (1)  & 37.5\% (3) \\
        Implementation \& Execution & 0.0\% (0)  & 31.2\% (5) & 12.5\% (1) \\
        \bottomrule
    \end{tabular}
\end{table}

\begin{table}[th]
    \centering
    \caption{Worst Case Modification Distribution by LLMs}
    \label{tab:mod_dist_worst_case_llms}
    \begin{tabular}{lccc}
        \toprule
        Modification Type & GPT-5 & Gemini-2.5-pro & Opus-4.1 \\
        \midrule
        Algorithmic Modification    & 25.0\% (4) & 68.8\% (11) & 37.5\% (3) \\
        Structural Modification     & 50.0\% (8) & 6.2\% (1)   & 12.5\% (1) \\
        Parameter Configuration     & 6.2\% (1)  & 12.5\% (2)  & 37.5\% (3) \\
        Implementation \& Execution & 18.8\% (3) & 12.5\% (2)  & 12.5\% (1) \\
        \bottomrule
    \end{tabular}
\end{table}

\begin{table}[th]
    \centering
    \caption{Best Case Modification Distribution by Agents}
    \label{tab:mod_dist_best_case_agents}
    \begin{tabular}{lccc}
        \toprule
        Modification Type & TheAIScientist & AIDE & Claude Code \\
        \midrule
        Algorithmic Modification & 75.0\% (12) & 83.3\% (10) & 37.5\% (3) \\
        Structural Modification  & 12.5\% (2)  & 0.0\% (0)   & 37.5\% (3) \\
        Data Processing          & 12.5\% (2)  & 16.7\% (2)  & 0.0\% (0)  \\
        Parameter Configuration  & 0.0\% (0)   & 0.0\% (0)   & 25.0\% (2) \\
        \bottomrule
    \end{tabular}
\end{table}

\begin{table}[!th]
    \centering
    \caption{Best Case Modification Distribution by LLMs}
    \label{tab:mod_dist_best_case_llms}
    \begin{tabular}{lccc}
        \toprule
        Modification Type & GPT-5 & Gemini-2.5-pro & Opus-4.1 \\
        \midrule
        Algorithmic Modification & 14 (100.0\%) & 8 (57.1\%) & 3 (37.5\%) \\
        Structural Modification  & 0 (0.0\%)    & 2 (14.3\%) & 3 (37.5\%) \\
        Data Processing          & 0 (0.0\%)    & 4 (28.6\%) & 0 (0.0\%)  \\
        Parameter Configuration  & 0 (0.0\%)    & 0 (0.0\%)  & 2 (25.0\%) \\
        \bottomrule
    \end{tabular}
\end{table}

\section{Protecting Evaluation Integrity}
A crucial aspect of our implementation involved protecting evaluation files from inadvertent modification by the agents. We employed agent-specific protection strategies: 
For TheAIScientist, we explicitly instructed the agent through prompting to avoid modifying evaluation files and implemented a systematic refresh mechanism that restores evaluation files before each evaluation cycle. 
For AIDE, protection is inherently ensured by its single-file modification constraint, which guarantees that when the target code and evaluation code are separated into different files, the evaluation files remain naturally protected.
For Claude Code, we implemented a two-pronged approach in which the evaluation files were first restricted to read and execute permissions before running the agent, and then the \texttt{--disallowedTools} argument was employed to explicitly prevent permission modification operations during execution.

\section{Additional Results}
\label{sec:app_experiments}

\begin{figure}[H]
    \centering
    \includegraphics[width=1\linewidth]{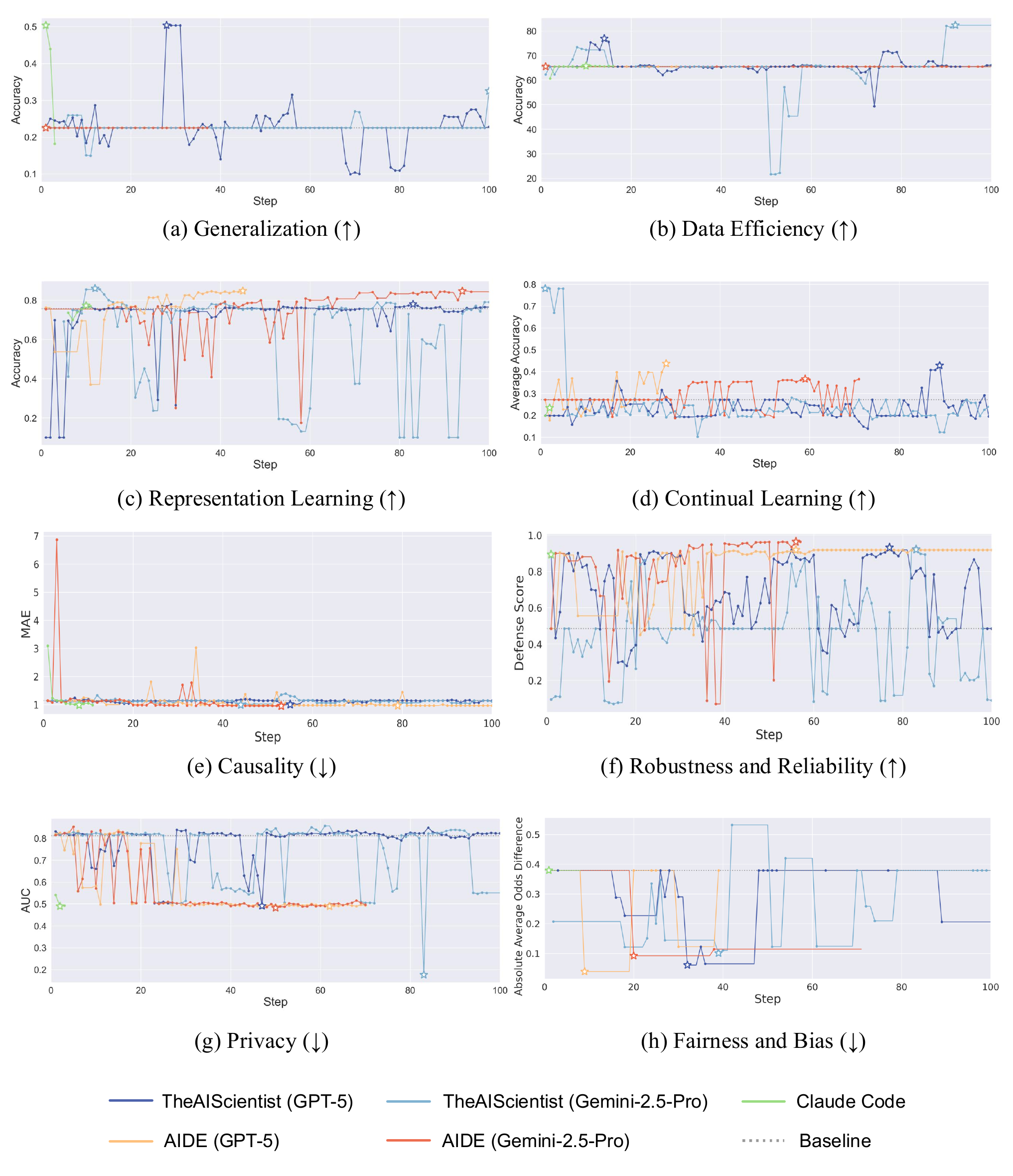}
    \caption{
    Agents' performance improvement curves across 8 tasks.
    }
    \label{fig:agents_research_iterations}
\end{figure}

This section provides a comprehensive comparative analysis of automated AI research systems, focusing on both their performance and operational efficiency. We evaluate TheAIScientist, AIDE, and Claude Code across eight core machine learning research problems, considering not only their final outcomes but also their operational behavior over time. 

Additional analysis results covers multiple key dimensions: detailed exploration diversity (Tab.~\ref{tab:diversity_comparsion}), operational reliability measured by step success rate (Tab.~\ref{tab:step_success_rate_comparison}) and step completion rate (Tab.~\ref{tab:step_completion_rate}), computational efficiency in terms of total token usage (Tab.~\ref{tab:total_tokens_comparison}) and step-wise token usage (Tab.~\ref{tab:step_tokens_comparison}), runtime efficiency in terms of total time (Tab.~\ref{tab:total_time_comparison_updated}) and step-wise time (Tab.~\ref{tab:step_time_comparison}). Together, these results provide detailed insights into the trade-offs between different agentic research designs, illustrating how they influence both the quality of research outcomes and the efficiency of resource utilization.

\begin{table}[H]
  \caption{ \textbf{Exploration Diversity Comparison of Best Performance Round.} G2.5-Pro denotes Gemini-2.5-Pro. Bold numbers indicate the best (highest) performance in each row.}
  \label{tab:diversity_comparsion}
  \centering
\renewcommand{\arraystretch}{0.9}
\begin{tabular}{l ccccc}
    \toprule[1.25pt]
    
    \multirow{2}{*}{\textbf{ML Problems}} & \multicolumn{2}{c}{\textbf{TheAIScientist$^*$}} & \multicolumn{2}{c}{\textbf{AIDE$^*$}} & \textbf{Claude Code} \\
    \cmidrule(lr){2-3} \cmidrule(lr){4-5} \cmidrule(lr){6-6}
      & GPT-5 & G2.5-Pro & GPT-5 & G2.5-Pro & Opus-4.1 \\

    \midrule

    Generalization      & 39.32 & 18.60 & 4.36  & 48.23 & 19.50 \\
    Data Eff.           & 28.18 & 20.19 & 18.03 & 10.92 & 10.08 \\
    Rep. Learn.         & 15.28 & 11.55 & 18.91 & 9.71  & 5.30 \\
    Cont. Learn.        & 18.86 & 24.83 & 36.72 & 18.80 & 2.71 \\
    Causality           & 24.03 & 21.37 & 38.61 & 11.65 & 6.43 \\
    Robust. \& Rel.     & 45.44 & 24.04 & 43.03 & 21.23 & 16.19 \\
    Privacy             & 29.66 & 21.17 & 33.16 & 13.67 & 5.97 \\
    Fair. \& Bias       & 26.88 & 25.14 & 35.97 & 13.10 & 3.84 \\

    \midrule
    Average $\pm$ Std   & 28.46$^{\pm 9.97}$ 
                        & 20.86$^{\pm 4.42}$ 
                        & 28.60$^{\pm 13.32}$ 
                        & 18.41$^{\pm 12.68}$ 
                        & 8.75$^{\pm 6.07}$ \\

    \bottomrule
    
\end{tabular}
\end{table}

\begin{table}[H]
  \caption{ \textbf{Step Success Rate Comparison.} 
G2.5-Pro denotes Gemini-2.5-Pro. Bold numbers indicate the best (highest) performance in each row.}
  \label{tab:step_success_rate_comparison}
  \centering
\renewcommand{\arraystretch}{0.9}
\begin{tabular}{l ccccc}
    \toprule[1.25pt]
    
    \multirow{2}{*}{\textbf{ML Problems}} & \multicolumn{2}{c}{\textbf{TheAIScientist$^*$}} & \multicolumn{2}{c}{\textbf{AIDE$^*$}} & \textbf{Claude Code} \\
    \cmidrule(lr){2-3} \cmidrule(lr){4-5} \cmidrule(lr){6-6}
      & GPT-5 & G2.5-Pro & GPT-5 & G2.5-Pro & Opus-4.1 \\

    \midrule

    Generalization      & 0.77 & 0.99 & 0.50 & 0.70 & \textbf{1.00} \\
    Data Eff.           & \textbf{0.96} & 0.65 & 0.23 & 0.29 & 0.94 \\
    Rep. Learn.         & 0.83 & \textbf{0.84} & 0.80 & 0.60 & 0.55 \\
    Cont. Learn.        & 0.96 & 0.94 & 0.82 & 0.92 & \textbf{1.00} \\
    Causality           & 0.98 & 0.89 & 0.88 & 0.98 & \textbf{1.00} \\
    Robust. \& Rel.     & 0.98 & 0.89 & 0.87 & 0.79 & \textbf{1.00} \\
    Privacy             & 0.97 & 0.95 & 0.87 & 0.96 & \textbf{1.00} \\
    Fair. \& Bias       & 0.18 & \textbf{0.21} & 0.18 & 0.04 & 0.13 \\

    \midrule
    Average $\pm$ Std   & \textbf{0.83}$^{\pm 0.27}$ 
                        & 0.80$^{\pm 0.26}$ 
                        & 0.64$^{\pm 0.30}$ 
                        & 0.66$^{\pm 0.34}$ 
                        & 0.83$^{\pm 0.32}$ \\

    \bottomrule
    
\end{tabular}
\end{table}

\begin{table}[H]
  \caption{ \textbf{Step Completion Rate (SCR) Comparison.} 
G2.5-Pro denotes Gemini-2.5-Pro. Bold numbers indicate the best (highest) performance in each row.}
  \label{tab:step_completion_rate}
  \centering
\renewcommand{\arraystretch}{0.9}
\begin{tabular}{l ccccc}
    \toprule[1.25pt]
    
    \multirow{2}{*}{\textbf{ML Problems}} & \multicolumn{2}{c}{\textbf{TheAIScientist$^*$}} & \multicolumn{2}{c}{\textbf{AIDE$^*$}} & \textbf{Claude Code} \\
    \cmidrule(lr){2-3} \cmidrule(lr){4-5} \cmidrule(lr){6-6}
      & GPT-5 & G2.5-Pro & GPT-5 & G2.5-Pro & Opus-4.1 \\

    \midrule

    Generalization      & \textbf{1.00} & \textbf{1.00} & 0.02 & 0.36 & 0.03 \\
    Data Eff.           & \textbf{1.00} & \textbf{1.00} & 0.30 & \textbf{1.00} & 0.16 \\
    Rep. Learn.         & \textbf{1.00} & \textbf{1.00} & 0.45 & \textbf{1.00} & 0.11 \\
    Cont. Learn.        & \textbf{1.00} & \textbf{1.00} & 0.28 & 0.70 & 0.02 \\
    Causality           & \textbf{1.00} & \textbf{1.00} & \textbf{1.00} & 0.52 & 0.11 \\
    Robust. \& Rel.     & \textbf{1.00} & \textbf{1.00} & \textbf{1.00} & 0.56 & 0.01 \\
    Privacy             & \textbf{1.00} & \textbf{1.00} & 0.70 & 0.69 & 0.03 \\
    Fair. \& Bias       & \textbf{1.00} & \textbf{1.00} & 0.39 & 0.7 & 0.08 \\

    \midrule
    Average $\pm$ Std   & \textbf{1.00}$^{\pm 0.00}$ 
                        & \textbf{1.00}$^{\pm 0.00}$ 
                        & 0.54$^{\pm 0.38}$ 
                        & 0.69$^{\pm 0.24}$ 
                        & 0.07$^{\pm 0.06}$ \\

    \bottomrule
    
\end{tabular}
\end{table}

\begin{table}[H]
  \caption{ \textbf{Total Token Consumption Comparison (in millions).} 
G2.5-Pro denotes Gemini-2.5-Pro. Bold numbers indicate the best (lowest) performance in each row. }
  \label{tab:total_tokens_comparison}
  \centering
\renewcommand{\arraystretch}{0.9}
\begin{tabular}{l ccc}
    \toprule[1.25pt]
    
    \multirow{2}{*}{\textbf{ML Problems}} & \multicolumn{2}{c}{\textbf{TheAIScientist$^*$}} & \textbf{Claude Code} \\
    \cmidrule(lr){2-3} \cmidrule(lr){4-4}
      & GPT-5 & G2.5-Pro & Opus-4.1 \\

    \midrule

    Generalization      & 9.34  & 9.88  & \textbf{5.98} \\
    Data Eff.           & 3.09  & \textbf{2.55}  & 7.98 \\
    Rep. Learn.         & 4.69  & \textbf{3.99}  & 8.89 \\
    Cont. Learn.        & 7.81  & 7.74  & \textbf{2.99} \\
    Causality           & 3.72	& \textbf{3.40}	& 10.86 \\
    Robust. \& Rel.     & 7.62	& \textbf{7.04}  & 13.36 \\
    Privacy             & 6.64  & \textbf{4.45}  & 13.88 \\
    Fair. \& Bias       & 5.41	& \textbf{4.58}	& 8.53 \\

    \midrule 
    Average $\pm$ Std   & 6.04$^{\pm 2.18}$ 
                        & \textbf{5.45}$^{\pm 2.51}$ 
                        & 9.06$^{\pm 3.64}$ \\

    \bottomrule
    
\end{tabular}
\end{table}

\begin{table}[H]
  \caption{ \textbf{Step Token Consumption Comparison (in millions).} 
G2.5-Pro denotes Gemini-2.5-Pro. Bold numbers indicate the best (lowest) performance in each row.}
  \label{tab:step_tokens_comparison}
  \centering
\renewcommand{\arraystretch}{0.9}
\begin{tabular}{l ccccc}
    \toprule[1.25pt]
    
    \multirow{2}{*}{\textbf{ML Problems}} & \multicolumn{2}{c}{\textbf{TheAIScientist$^*$}} & \textbf{Claude Code} \\
    \cmidrule(lr){2-3} \cmidrule(lr){4-4}
      & GPT-5 & G2.5-Pro & Opus-4.1 \\

    \midrule

    Generalization      & \textbf{0.09}  & 0.10   & 1.99 \\
    Data Eff.           & \textbf{0.03}  & \textbf{0.03}  & 0.50 \\
    Rep. Learn.         & 0.05  & \textbf{0.04}  & 0.81 \\
    Cont. Learn.        & \textbf{0.08}  & \textbf{0.08}  & 1.49 \\
    Causality           & 0.04	& \textbf{0.03}	& 0.99 \\
    Robust. \& Rel.     & 0.08	& \textbf{0.07}	& 13.36 \\
    Privacy             & 0.07  & \textbf{0.04}  & 4.63 \\
    Fair. \& Bias       & \textbf{0.05}	& \textbf{0.05}	& 1.07 \\

    \midrule
    Average $\pm$ Std   & 0.06$^{\pm 0.02}$ 
                        & \textbf{0.05}$^{\pm 0.03}$ 
                        & 3.10$^{\pm 4.34}$ \\

    \bottomrule
    
\end{tabular}
\end{table}

\begin{table}[H]
  \caption{ \textbf{Total Time Consumption Comparison (in hours).} 
G2.5-Pro denotes Gemini-2.5-Pro. Bold numbers indicate the best (lowest) performance in each row.}
  \label{tab:total_time_comparison_updated}
  \centering
\renewcommand{\arraystretch}{0.9}
\begin{tabular}{l ccccc}
    \toprule[1.25pt]
    
    \multirow{2}{*}{\textbf{ML Problems}} & \multicolumn{2}{c}{\textbf{TheAIScientist$^*$}} & \multicolumn{2}{c}{\textbf{AIDE$^*$}} & \textbf{Claude Code} \\
    \cmidrule(lr){2-3} \cmidrule(lr){4-5} \cmidrule(lr){6-6}
      & GPT-5 & G2.5-Pro & GPT-5 & G2.5-Pro & Opus-4.1 \\

    \midrule
    
    Generalization      & 9.91  & 17.42 & \textbf{0.20} & 6.07 & 0.81 \\
    Data Eff.           & 4.47  & 4.33  & 1.02  & 2.40  & \textbf{0.47} \\
    Rep. Learn.         & 7.50  & 23.96 & 4.45  & 14.49 & \textbf{5.56} \\
    Cont. Learn.        & 14.68 & 15.17 & 3.70  & 7.08  & \textbf{0.28} \\
    Causality           & 4.97 & 3.51 & 4.50 & 3.43 & \textbf{0.40} \\
    Robust. \& Rel.     & 19.40 & 6.60 & 22.02 & 17.84 & \textbf{0.32} \\
    Privacy             & 9.24  & 20.03 & 2.43  & 21.04 & \textbf{1.29} \\
    Fair. \& Bias       & 5.20 & 2.87 & 5.32 & 3.60 & \textbf{0.30} \\

    \midrule
    Average $\pm$ Std   & 9.42$^{\pm 5.25}$ 
                        & 11.74$^{\pm 8.37}$ 
                        & 5.46$^{\pm 6.92}$ 
                        & 9.49$^{\pm 7.24}$ 
                        & \textbf{1.18}$^{\pm 1.80}$ \\

    \bottomrule
    
\end{tabular}
\end{table}

\begin{table}[H]
  \caption{ \textbf{Step Time Consumption Comparison (in minutes).} 
G2.5-Pro denotes Gemini-2.5-Pro. Bold numbers indicate the best (lowest) performance in each row.}
  \label{tab:step_time_comparison}
  \centering
\renewcommand{\arraystretch}{0.9}
\begin{tabular}{l ccccc}
    \toprule[1.25pt]
    
    \multirow{2}{*}{\textbf{ML Problems}} & \multicolumn{2}{c}{\textbf{TheAIScientist$^*$}} & \multicolumn{2}{c}{\textbf{AIDE$^*$}} & \textbf{Claude Code} \\
    \cmidrule(lr){2-3} \cmidrule(lr){4-5} \cmidrule(lr){6-6}
      & GPT-5 & G2.5-Pro & GPT-5 & G2.5-Pro & Opus-4.1 \\

    \midrule

    Generalization      & \textbf{5.95} & 10.43 & 5.97 & 9.83 & 16.13 \\
    Data Eff.           & 2.68 & 2.60 & 2.03 & \textbf{1.43} & 1.77 \\
    Rep. Learn.         & 33.28 & 43.15 & 37.92 & \textbf{22.87} & 30.30 \\
    Cont. Learn.        & 8.80 & 9.08 & 7.93 & \textbf{5.98} & 8.33 \\
    Causality           & 2.98 & \textbf{2.10} & 2.68 & 3.88 & 2.20 \\
    Robust. \& Rel.     & \textbf{11.63} & 18.37 & 13.08 & 18.78 & 19.20 \\
    Privacy             & 19.93 & 26.42 & 22.63 & \textbf{18.03} & 25.87 \\
    Fair. \& Bias       & 3.12 & 1.72 & 8.18 & \textbf{3.05} & 2.25 \\

    \midrule
    Average $\pm$ Std   & 11.05$^{\pm 10.71}$ 
                        & 14.23$^{\pm 14.55}$ 
                        & 12.55$^{\pm 12.17}$ 
                        & \textbf{10.48}$^{\pm 8.29}$ 
                        & 13.26$^{\pm 11.29}$ \\

    \bottomrule
    
\end{tabular}
\end{table}

\section{Prompts Details}
This section presents the detailed prompt specifications that form the foundation of our automatic research agent framework. The prompts serve as the primary interface between human researchers and AI agents, translating high-level research objectives into actionable instructions that can guide systematic scientific inquiry across diverse machine learning domains.
The prompt design philosophy centers on creating a structured yet flexible research environment that balances autonomy with scientific rigor. Rather than providing overly prescriptive instructions that limit creative exploration, these prompts establish clear boundaries, evaluation criteria, and operational constraints while encouraging the agent to develop and test novel hypotheses within established research paradigms.

This section is organized into two complementary components. First, we present the Task Description Prompts that define specific research challenges across 8 fundamental areas of machine learning, each grounded in established benchmarks and methodologies. These prompts simulate realistic research scenarios where an AI agent must navigate complex technical requirements while pursuing meaningful improvements to existing methods.

Second, we detail the Automatic Research Agent Framework that governs how agents interact with research codebases to conduct iterative experimentation. This operational framework transforms the conceptual research challenges into executable workflows, ensuring that agent behavior follows sound scientific methodology while maintaining reproducibility and experimental integrity.
Together, these prompt specifications create a comprehensive research environment where automatic agents can contribute meaningfully to advancing machine learning across multiple disciplines, providing both the research contexts and the methodological framework necessary for systematic scientific progress.

\paragraph{Task Description Prompts}
The following task descriptions establish comprehensive research contexts spanning 8 fundamental areas of machine learning. Each prompt follows a structured format that defines: 1) the researcher's identity and expertise, 2) the specific technical setup including datasets and baseline methods, 3) clear optimization objectives and constraints, and 4) fairness criteria to ensure meaningful comparisons.

These prompts span a wide range of machine learning challenges (e.g., generalization, data efficiency, privacy, fairness, and robustness). Each task is anchored in established benchmarks and frameworks, such as DomainBed for domain generalization, EasyFSL for few-shot learning, and AIF360 for fairness-aware learning, providing realistic experimental environments that closely mirror actual research workflows.

The prompts are carefully crafted to encourage both incremental improvements to existing methods and the exploration of novel algorithmic approaches, while maintaining scientific rigor through controlled experimental conditions. This design enables systematic evaluation of how automatic agents can contribute to advancing the state-of-the-art across multiple ML disciplines simultaneously.

\vspace{20mm}

\subparagraph{Generalization} \

\begin{tcolorbox}[colback=gray!10, colframe=black!50, boxrule=0.5pt, arc=2mm]
\textbf{System:} You are an ambitious AI PhD student focused on improving the generalization performance of machine learning methods using the DomainBed benchmark.\\

\textbf{Task Description:} You are working with DomainBed's ERM (Empirical Risk Minimization) method as the baseline on ColoredMNIST to evaluate generalization under distribution shifts. Your goal is to enhance test-time domain generalization accuracy beyond standard ERM. You should improve the algorithm based on ERM, but you may also propose entirely new algorithms if they can better support cross-domain generalization. You are also allowed to refine the backbone model, as long as your modifications are fair compared to the original architecture. The priority is to improve the average accuracy on unseen test domains while maintaining accuracy on in-domain tests, along with ensuring efficiency and low complexity.
\end{tcolorbox}

\subparagraph{Data Efficiency} \

\begin{tcolorbox}[colback=gray!10, colframe=black!50, boxrule=0.5pt, arc=2mm]
\textbf{System:} You are an ambitious AI PhD student focused on data-efficient learning, specializing in few-shot learning and meta-learning.\\

\textbf{Task Description:} You are working with the EasyFSL framework to enhance the FewShotClassifier on the Mini-ImageNet dataset. The Mini-ImageNet dataset presents a challenging few-shot learning scenario due to its fine-grained inter-class similarities and limited training examples per class. Your goal is to improve the classifier’s ability to generalize to novel classes.
\end{tcolorbox}

\subparagraph{Representation Learning} \

\begin{tcolorbox}[colback=gray!10, colframe=black!50, boxrule=0.5pt, arc=2mm]
\textbf{System:} You are an ambitious AI PhD student focused on improving representation learning on CIFAR-10 using the Lightly self-supervised learning framework.\\

\textbf{Task Description:} You are working with Lightly’s MoCo baseline on CIFAR-10, evaluated strictly by linear probing Top-1 accuracy. Your goal is to improve representation learning at pretrain stage to improve linear-probe accuracy on the CIFAR-10 test set beyond standard MoCo as much as you can under the same compute and data (no external data). You may modify MoCo or propose new self-supervised methods if they can yield better representations, as long as your modifications are fair compared to the original architecture. You are also allowed to refine the ResNet-18 backbone as long as parameter count and FLOPs remain comparable to the baseline. Pretrain on the CIFAR-10 train split without labels, fit the linear classifier on the same train split, and report Top-1 on the test split with priority on improving representation learning performance.
\end{tcolorbox}

\subparagraph{Continual Learning} \

\begin{tcolorbox}[colback=gray!10, colframe=black!50, boxrule=0.5pt, arc=2mm]
\textbf{System:} You are an ambitious AI PhD student focused on improving continual learning based on Synaptic Intelligence (SI) on splitMNIST under the class-incremental scenario.\\

\textbf{Task Description:} You are working with the Continual-Learning repo’s SI baseline to improve accuracy and reduce forgetting on splitMNIST (5 tasks × 2 classes, single-head over 10 classes). Your goal is to improve average accuracy over all 5 contexts on splitMNIST without unfair model size or compute advantages. You should improve SI method, but are also allowed to add lightweight fair components , or propose new methods, as long as your modifications are fair (stay within fairness computation budgets). The priority is to improve the average accuracy.
\end{tcolorbox}

\subparagraph{Causality} \

\begin{tcolorbox}[colback=gray!10, colframe=black!50, boxrule=0.5pt, arc=2mm]
\textbf{System:} You are an ambitious AI PhD student focused on advancing machine learning for causal inference, reasoning, and interpretable modeling.\\

\textbf{Task Description:} You are working with the Dragonnet framework to estimate individual treatment effects (ITEs) in IHDP. The Infant Health and Development Program (IHDP) is a semi-synthetic benchmark dataset commonly used in causal inference, built from a real randomized trial with simulated treatment effects to enable ground-truth evaluation. Your goal is to improve the precision of treatment effect estimation across IHDP.
\end{tcolorbox}

\subparagraph{Robustness and Reliability} \

\begin{tcolorbox}[colback=gray!10, colframe=black!50, boxrule=0.5pt, arc=2mm]
\textbf{System:} You are an ambitious AI PhD student focused on improving robust learning under data poisoning and privacy constraints.\\

\textbf{Task Description:} You are given the Adversarial Robustness Toolbox (ART) codebase with a focus on the \texttt{dp\_instahide} defense. \texttt{dp\_instahide} mixes inputs with public data and applies differential privacy noise to hinder inversion and poisoning. While designed for privacy-preserving training, its structure offers headroom to harden against both clean-label and trigger/backdoor poisons. Your goal is to improve defense performance against diverse poisoning attacks while maintaining high clean accuracy. You may tune \texttt{dp\_instahide}, compose it with other defenses, or propose a new method if it outperforms baselines.

\end{tcolorbox}

\subparagraph{Privacy} \ 

\begin{tcolorbox}[colback=gray!10, colframe=black!50, boxrule=0.5pt, arc=2mm]
\textbf{System:} You are an ambitious AI PhD student focused on improving model privacy and security against membership inference attacks.\\

\textbf{Task Description:} You are working with PrivacyMeter’s MIA (for information leakage through training points) and Robust MIA (RMIA, which refines the Likelihood Ratio Test with a tighter null hypothesis and leverages reference models and population data) to evaluate and reduce the model’s privacy risk. Your goal is to drive the auditor’s AUC toward 0.5 and keep TPR@0.1\%FPR and TPR@0.0\%FPR near zero while preserving task accuracy. Focus only on defense-side strategies rather than modifying the attack algorithms.
\end{tcolorbox}

\subparagraph{Fairness and Bias} \ 

\begin{tcolorbox}[colback=gray!10, colframe=black!50, boxrule=0.5pt, arc=2mm]
\textbf{System:} You are an ambitious AI PhD student focused on improving fairness-aware learning with AIF360’s Adversarial Debiasing on the COMPAS dataset.\\

\textbf{Task Description:} You are working with AIF360’s Adversarial Debiasing (classifier–adversary) as the baseline on the COMPAS dataset to evaluate the fairness–accuracy trade-off. Your goal is to minimize absolute Average Odds Difference toward parity (=0) while maintaining or improving Balanced Accuracy on held-out test splits and across protected subgroups (e.g., sex/race). You should enhance the baseline Adversarial Debiasing algorithm, but you may also propose entirely new fairness methods if they better support reduced absolute Average Odds Difference without sacrificing Balanced Accuracy. You are allowed to refine the classifier and adversary networks and the training pipeline, provided comparisons remain fair to the original setup (similar capacity, training budget, and data access). The priority is minimizing absolute Average Odds Difference while preserving or improving Balanced Accuracy.
\end{tcolorbox}

\paragraph{Prompting Claude Code as an Automatic Research Agent}
\label{para:claude_code_prompt}
The following comprehensive prompt specification defines how an AI agent operates within established research codebases to achieve meaningful scientific progress. Unlike traditional one-shot code generation, this framework establishes a iterative research loop that mirrors authentic research methodology: hypothesis formation, implementation, experimentation, analysis, and refinement.

\begin{tcolorbox}[colback=purple!5!white,colframe=purple!75!black,title=Your Role]
You are an autonomous coding agent that:
\begin{itemize}
\item understands the task,
\item proposes a concrete idea/plan/solution,
\item edits the code (respecting read-only constraints),
\item executes a fixed command list,
\item handles errors by diagnosing and fixing them,
\item records each step's modifications and results, and
\item iterates until the iteration limit is reached.
\end{itemize}
\end{tcolorbox}

\begin{tcolorbox}[colback=violet!5!white,colframe=violet!75!black,title=Repository Access]
You are given starter files, \texttt{STARTER\_FILE\_PATHS}, and may read other files as needed to complete the task.

\textbf{Hard constraint:} Do not modify any file whose path matches \texttt{READONLY\_PATHS}. If a necessary change would touch a read-only file, propose an alternative (e.g., wrapper, config flag, adapter module) instead.
\end{tcolorbox}

\begin{tcolorbox}[colback=orange!5!white,colframe=orange!75!black,title=Loop Initialization]
\textbf{Initialize:}
\begin{itemize}
\item count = 0
\item Record a snapshot/baseline of the original code (the repository state before any of your edits). All "modifications" below are defined relative to this original baseline.
\item Read original baseline results for reference. The results are provided in \texttt{ORIGINAL\_BASELINE\_RESULTS\_PATH}.
\end{itemize}
\end{tcolorbox}

\begin{tcolorbox}[colback=olive!5!white,colframe=olive!75!black,title=Step 1-3: Understanding and Planning]
\textbf{Step 1 — Understand the task}
\begin{itemize}
\item Read the repo and \texttt{TASK\_DESCRIPTION}.
\item If helpful, quickly inventory key entry points, configs, data paths, and any training/eval scripts.
\end{itemize}

\textbf{Step 2 — Generate a plan}
\begin{itemize}
\item Produce a brief idea/plan/solution describing what you will change and why.
\end{itemize}

\textbf{Step 3 — Modify the code}
\begin{itemize}
\item Implement your plan with minimal, focused edits.
\item Respect \texttt{READONLY\_PATHS} at all times (no renames, moves, or edits under those paths).
\item Keep changes atomic and well-commented.
\end{itemize}
\end{tcolorbox}

\begin{tcolorbox}[colback=red!5!white,colframe=red!75!black,title=Step 4-6: Execution and Error Handling]
\textbf{Step 4 — Execute commands}
\begin{itemize}
\item Run every command in \texttt{COMMAND\_LIST} sequentially.
\item Capture stdout/stderr and exit codes for each command.
\item After the command list completes (whether fully or interrupted by an error), do: count += 1.
\end{itemize}

\textbf{Step 5 — Error handling}
If any command raised an exception or returned a non-zero exit code:
\begin{itemize}
\item Diagnose the exception concisely (root cause + where it occurred).
\item Propose a specific fix.
\item Apply the fix by editing the code (still respecting \texttt{READONLY\_PATHS}).
\item Proceed to the next iteration (go back to Step 4 for another execution after modifications).
\end{itemize}

\textbf{Step 6 — Iteration limit}
\begin{itemize}
\item If count $>$ \texttt{MAX\_ITERS}, stop and produce a final summary.
\end{itemize}
\end{tcolorbox}

\begin{tcolorbox}[colback=cyan!5!white,colframe=cyan!75!black,title=Step 7-10: Backup and Results Management]
\textbf{Step 7 — Per-step backup (always)}
For each iteration (each time you execute the command list), create a directory:
\texttt{./\_agent\_runs/step\_\{COUNT\}/}

Store in that directory:
\begin{itemize}
\item \texttt{modified/} → only the files that differ from the original baseline (preserve their relative paths).
\item \texttt{logs/} → command outputs (one file per command, including exit codes).
\end{itemize}

\textbf{Step 8 — Successful run artifacts}
If all commands in \texttt{COMMAND\_LIST} completed successfully:
\begin{itemize}
\item In \texttt{./\_agent\_runs/step\_\{COUNT\}/results/}, \\copy: \texttt{\$\{RESULT\_DIR\}/final\_info.json}
\item Confirm that \texttt{final\_info.json} exists in the step results.
\end{itemize}

\textbf{Step 9 — Read \& reset results directory}
\begin{itemize}
\item Read and summarize the key contents of \texttt{final\_info.json} for guidance.
\item Then delete the entire \texttt{RESULT\_DIR} to avoid conflicts with future iterations.
\end{itemize}

\textbf{Step 10 — Improve the plan}
\begin{itemize}
\item Based on the results, your current idea/plan/solution, and the code modifications so far: Generate a new idea/plan/solution to further improve the outcome.
\item Continue the loop, unless the iteration limit has been reached.
\end{itemize}
\end{tcolorbox}

\begin{tcolorbox}[colback=yellow!5!white,colframe=yellow!75!black,title=Additional Rules \& Conventions]
\textbf{Command semantics:} Treat any non-zero exit code as an error. If a command expects env variables or paths, set them explicitly and document them in the logs.

\textbf{Diff discipline:} When storing modified/ files for a step, include only files that differ from the original baseline (not from the previous step).

\textbf{Execution discipline:}
\begin{itemize}
\item Execute \texttt{COMMAND\_LIST} verbatim: do not change the order, arguments, flags, prefixes (e.g., env vars), or wrap the commands.
\item Resolve failures by modifying code or configuration (outside \texttt{READONLY\_PATHS}) instead of altering the commands.
\end{itemize}

\textbf{Other rules:}
\begin{itemize}
\item \textbf{Atomic changes:} Prefer small, testable edits.
\item \textbf{Evaluation integrity:} Never alter evaluation logic, datasets, or scripts inside \texttt{READONLY\_PATHS}.
\item \textbf{Idempotence:} If a prior step succeeded, avoid regressing it.
\end{itemize}

\textbf{Important constraint:} Under no circumstances may you halt while the current step count $\leq$ \texttt{MAX\_ITERS}; you must continue the modify $\rightarrow$ execute \texttt{COMMAND\_LIST} $\rightarrow$ diagnose/fix loop.
\end{tcolorbox}

\begin{tcolorbox}[colback=gray!5!white,colframe=gray!75!black,title=Optimization Directions]
\textbf{Optimization directions:} The global setting, \texttt{OPTIMIZATION\_DIRECTION}, defines the default direction for all metrics and accepts either "higher" or "lower". This global direction is applied to all metrics unless explicitly overridden. For more granular control, \texttt{PER\_METRIC\_DIRECTION} provides a way to override the global setting by specifying a mapping of individual metric names to their desired optimization direction.

\textbf{Optimization goal filtering:} If optimization target metrics (\texttt{TARGET\_METRICS}) and dataset (\texttt{TARGET\_DATASETS}) names are provided, treat them as strict optimization goals: focus exclusively on improving the specified metrics on the specified datasets, and ignore all other metrics and datasets.

\textbf{Runtime Logging Rule:} You must log the start and end time of the entire process. Record wall-clock timestamps when you begin and exit the loop. Save timestamps to \texttt{./\_agent\_runs/process\_time\_log.txt} with format:
\begin{verbatim}
start_time: 2025-08-14 13:04:22
end_time:   2025-08-14 14:37:55
duration_seconds: 5573
\end{verbatim}
\end{tcolorbox}

\begin{tcolorbox}[colback=red!10!white,colframe=red!75!black,title=CRITICAL EXECUTION REQUIREMENT,fontupper=\bfseries]
For ALL commands in \texttt{COMMAND\_LIST} that are expected to take more than 5 minutes (especially training commands with epochs $>$ 10), you MUST:

\begin{enumerate}
\item ALWAYS use \texttt{run\_in\_background=True} when executing these commands with the Bash tool
\item Monitor the background process using BashOutput tool with the returned \texttt{bash\_id}
\item Wait for completion by periodically checking BashOutput until the process finishes
\item Only proceed to the next command after confirming the previous background process has completed successfully
\end{enumerate}

\textbf{Example execution pattern:}
\begin{verbatim}
result = Bash(
    command="python main.py ... --num_epochs 100 ...",
    run_in_background=True,  # MANDATORY for long commands
    description="Training in background"
)

shell_id = result.split(
    "Background shell started with ID: "
)[1].split("\n")[0]

# Monitor until completion
while True:
    output = BashOutput(bash_id=shell_id)
    if "still running" not in output:
        break
    time.sleep(30)  # Check every 30 seconds

\end{verbatim}

\textbf{FAILURE TO USE BACKGROUND EXECUTION FOR LONG-RUNNING COMMANDS WILL BE CONSIDERED A CRITICAL ERROR.}

Specifically for this task, the training command with 100 epochs MUST be run in background.
\end{tcolorbox}

\begin{tcolorbox}[colback=teal!5!white,colframe=teal!75!black,title=Reporting ]
After each iteration, output a short report with:
\begin{itemize}
\item count
\item Idea/Plan/Solution (current) — 3–6 bullet points
\item Changes Made — list of files edited with 1-line rationale each
\item Command Results — success/failure per command with exit code
\item If success: brief summary of \texttt{final\_info.json} key metrics
\item Next Steps — what you'll try next (or stop if count $>$ \texttt{MAX\_ITERS})
\end{itemize}
\end{tcolorbox}

\begin{tcolorbox}[colback=brown!5!white,colframe=brown!75!black,title=Begin with Template Variables]
\texttt{TASK\_DESCRIPTION:} ``"

\texttt{STARTER\_FILE\_PATHS:} [ ]

\texttt{READONLY\_PATHS:} [``"]

\texttt{ORIGINAL\_BASELINE\_RESULTS\_PATH:} ``"

\texttt{TARGET\_METRICS:} [``"] 

\texttt{TARGET\_DATASETS:} [``"]

\texttt{OPTIMIZATION\_DIRECTION:} ``"

\texttt{PER\_METRIC\_DIRECTION} = \{ \}

\texttt{COMMAND\_LIST:} [ ]

\texttt{MAX\_ITERS:} 

\texttt{RESULT\_DIR:} ``"
\end{tcolorbox}


\end{document}